\documentclass[conference]{IEEEtran}
\IEEEoverridecommandlockouts
\usepackage{cite}
\usepackage{amsmath,amssymb,amsfonts}
\usepackage{algorithmic}
\usepackage{graphicx}
\usepackage{textcomp}
\usepackage{booktabs}
\usepackage{multirow}
\usepackage{xcolor}
\def\BibTeX{{\rm B\kern-.05em{\sc i\kern-.025em b}\kern-.08em
    T\kern-.1667em\lower.7ex\hbox{E}\kern-.125emX}}
\begin{document}

\title{TPLogAD: Unsupervised Log Anomaly Detection Based on Event Templates and Key Parameters}

\author{\IEEEauthorblockN{1\textsuperscript{st} Jiawei Lu}
\IEEEauthorblockA{\textit{School of Computer Science and Technology} \\
\textit{Fudan University}\\
Shanghai, China \\
jwlu22@m.fudan.edu.cn}
\and
\IEEEauthorblockN{2\textsuperscript{nd} Chengrong Wu}
\IEEEauthorblockA{\textit{School of Computer Science and Technology} \\
\textit{Fudan University}\\
Shanghai, China \\
cwu@fudan.edu.cn}}

\maketitle

\begin{abstract}
Log-system is an important mechanism for recording the runtime status and events of Web service systems, and anomaly detection in logs is an effective method of detecting problems. However, manual anomaly detection in logs is inefficient, error-prone, and unrealistic. Existing log anomaly detection methods either use the indexes of event templates, or form vectors by embedding the fixed string part of the template as a sentence, or use time parameters for sequence analysis. However, log entries often contain features and semantic information that cannot be fully represented by these methods, resulting in missed and false alarms. In this paper, we propose TPLogAD, a universal unsupervised method for analyzing unstructured logs, which performs anomaly detection based on event templates and key parameters. The itemplate2vec and para2vec included in TPLogAD are two efficient and easy-to-implement semantic representation methods for logs, detecting anomalies in event templates and parameters respectively, which has not been achieved in previous work. Additionally, TPLogAD can avoid the interference of log diversity and dynamics on anomaly detection. Our experiments on four public log datasets show that TPLogAD outperforms existing log anomaly detection methods.
\end{abstract}


\section{Introduction}
Log-system is a crucial element of Web service systems as they record the operational status, behaviors, and events of these systems, reflecting their performance and security\cite{b1}. And logs can assist administrators in monitoring, debugging, and maintaining Web service systems, enhancing their reliability and availability\cite{b2}.

However, the scale and complexity of Web service systems have led to an explosive growth in the types, quantities, and contents of logs. This growth presents significant challenges for log analysis and processing, making manual log detection unfeasible\cite{b3}. Moreover, different modules of Web service systems may use various log formats, languages, and standards, and their operating environments and requirements related to logs may change at any time. Furthermore, logs often contain a significant amount of unstructured data, including free text, symbols, and parameters. This diversity and dynamic nature of logs can make analysis and processing difficult.

To effectively analyze logs and improve the quality of Web service systems, automated log anomaly detection has become an important technology\cite{b4}. Log anomaly detection usually begins with log parsing, which extracts fixed strings of similar events in log entries into event templates and variable strings into parameters. Then the content and structure of logs are analyzed to identify abnormal states and events that do not conform to normal states and events implied in logs. This helps to discover potential errors, failures, and attacks in Web service systems, thereby enhancing their stability and security.

According to the features of logs, log anomalies can be classified into two categories: sequence anomalies and parameter anomalies. Sequence anomalies refer to inconsistencies between the order of events corresponding to log entries and the normal order, indicating errors in the process. Parameter anomalies are significant differences between the value of a parameter of a certain operation behavior or state in the log entries, and the value of that parameter under normal situation. They usually indicate misoperations or fluctuations in the performance of some components. Current log anomaly detection methods can be classified into two categories: those based on event templates\cite{b5,b6,b7,b8} and those based on parameters correlation analysis\cite{b9,b10,b11,b12}. The former extracts event template features from logs, converts them into vectors, and then uses the distance or similarity between vectors to identify sequence anomalies. The latter analyzes the correlation and dependency relationship between events in logs by utilizing time parameters, constructs directed graphs, and then uses the structure and properties of the graph to identify parameter anomalies. 

Existing methods for templates typically use template indexes or treat the fixed string of the template as a sentence, forming vectors with weak semantics based on synonyms and antonyms. However, this approach ignores the richer semantic and structural information contained in the parameters of the log entries. Similarly, for parameters, existing methods generally conduct correlation analysis based on time parameters that are generated with the same format, but this ignores more valuable parameters such as user identification, system status, and resource identification. Therefore, existing log anomaly detection methods have the following three challenges.

\begin{itemize}
    \item \textit{Semantic loss}. The template indexes and the synonyms and antonyms in the fixed string of the template cannot fully represent the semantic information of the template. In many cases, words do not have a clear semantic relationship and must be understood based on context and situation. For instance, the words $login$ and $register$ are not synonyms or antonyms, but they are related. Neglecting this correlation may lead to misjudgment.
    \item \textit{Parameter abandonment}. Most existing methods only consider event templates and time parameters generated by the log entries\cite{b13}. Furthermore, log entries for different types of events have various parameter types, which are difficult to analyze with uniform processing method by the detection module. This is why most existing methods exclude parameters when dealing with logs of mixed-type events. As a result, existing methods cannot comprehensively identify abnormal logs by combining event templates and parameters.
    \item \textit{Difficulty in adapting to the diversity and dynamics of logs}. Web service systems may use various log formats and standards, and their operating environments and requirements may change frequently. For instance, events in the logs may have different dependencies due to changes in user identification, updates of event types, and other factors. Failure or delay of anomaly detection may occur if the anomaly detection is not updated timely to adapt to these changes in the logs.
\end{itemize}

To tackle the challenges mentioned above, we present TPLogAD, a deep learning framework that uses event templates and key parameters for unsupervised anomaly detection in unstructured log streams. The innovative points of TPLogAD compared to existing methods are as follows.

\begin{itemize}
    \item We propose itemplate2vec, to extract semantic information more effectively from event templates. Bert\cite{b14} is used to learn semantic relationship between words in logs. Template vectors are constructed through semantic similarity, which accurately extract semantic information from log templates for anomaly detection.
    \item Inspired by time2vec\cite{b15}, we design an efficient parameter representation method, para2vec, to accurately extract semantic information from parameters in logs. Clearly, para2vec uses an adaptive clustering algorithm to select key parameters that can characterize main features from a large number of parameters in logs. Then we design vector encoding methods for five parameter types: time, user identification, value, status, and resource identification. These five parameter types are commonly found in logs during real-world log researches.\cite{b1,b13}.
    \item We design a method for updating templates and parameters to handle the diversity and dynamics of logs, without requiring manual feedback from administrators.
    \item We propose TPLogAD, an framework utilizing BiLSTM and attention mechanism to learn semantic and association information in event templates and parameters. To the best of our knowledge, this is the first log anomaly detection that considers both event templates and parameters jointly.
\end{itemize}

We evaluate TPLogAD with existing log anomaly detection methods on four public benchmark datasets. The results show that TPLogAD outperforms state-of-the-art log anomaly detection methods.
\begin{figure}
    \centering
    \includegraphics[width=1\linewidth]{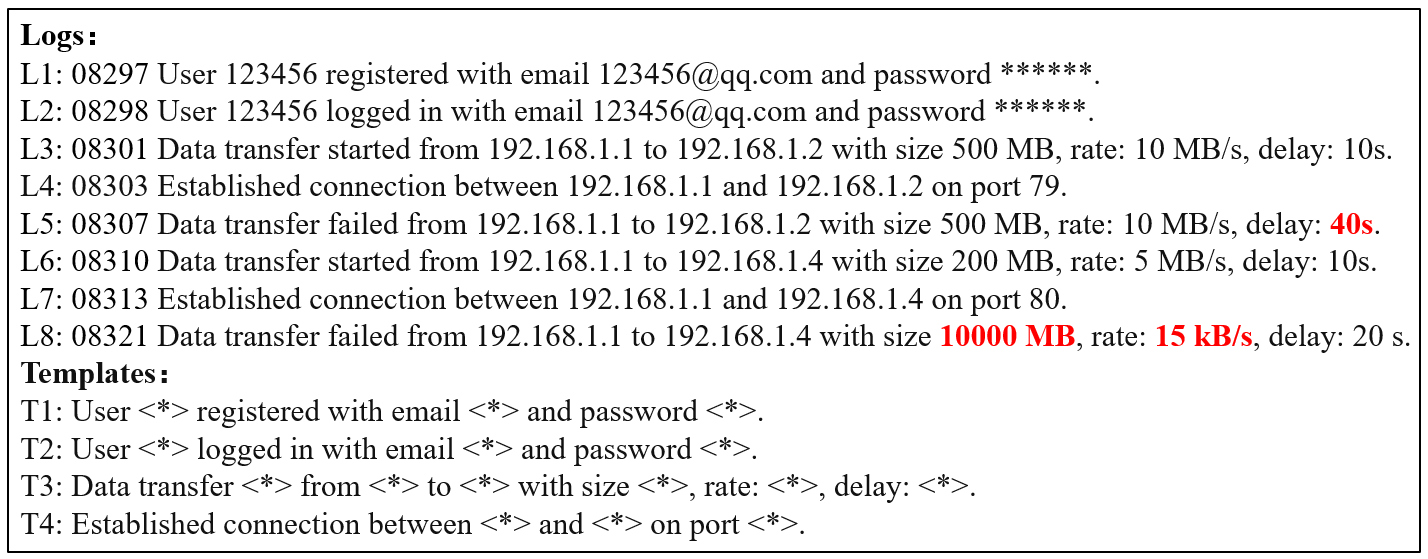}
    \caption{Logs and templates}
    \label{fig:log-example}
\end{figure}
\section{BackGround}

Most log parsing methods view log entries as a combination of event templates and parameters. Event templates are fixed strings that predefined by developers, and describe the type of events recorded by log entries. Parameters are variables used by the program to provide details of a log entry. However, when using templates formed by log parsing, parameters are replaced with $<*>$, so that templates no longer contain parameter content. For instance, in Fig.~\ref{fig:log-example}, the event template of log entry $L4$ is $T4$, which indicates that the system has established a data transmission connection. The parameters are “192.168.1.1”, “192.168.1.2”, and “79”, representing the IP address and port number of the server.

To analyze and extract information from logs, it is necessary to convert them into structured data. This process is known as log parsing\cite{b16,b17}. As illustrated in Fig.~\ref{fig:log-example}, log entries $L5$ and $L8$ both belong to template $T3$, indicating a failure in the system's data transmission. The cause of the failure is highlighted in bold red in the parameters. $L5$ is attributed to excessive delay, while $L8$ is caused by the transmission rate being slow. However, this information is reflected in parameters. Failure to consider these parameters will make it impossible to identify the root cause of the anomaly. To achieve efficient log anomaly detection, we utilize Drain3 for log parsing\cite{b18}. Drain3 can simultaneously separate templates and parameters in logs, and has high accuracy in template extraction\cite{b19}. It can also perform online parsing when new logs appear.

Log anomaly detection is a crucial aspect of log analysis as it enables the identification of system failures, attacks and other issues. In recent years, there has been significant progress in methods of log anomaly detection, which can be broadly classified into two categories: 1)based on event templates; 2)based on parameters correlation analysis.

Methods based on event templates involve parsing logs into structured event templates, constructing features using the semantic information in templates, and then using these features to detect log anomalies. For instance, LogAnomaly\cite{b5} uses LSTM to learn the semantic relationship of synonyms and antonyms. DeepLog\cite{b8} utilizes template indexes, and declares anomalies based on whether the predicted next log matches the actual log. However, LogRobust\cite{b6} uses adversarial training to enhance the robustness of log anomaly detection. Logsy\cite{b7} uses multi-task learning and self-attention mechanism to learn the features, and employs density estimation to detect log anomalies. LogBERT\cite{b20} utilizes Bert to acquire the semantic representation of the complete logs.

Methods based on parameters correlation analysis typically construct directed graphs for events according to the time parameters in logs, and then detect log anomalies using the association information on the graph. For example, Log2vec\cite{b9} builds a heterogeneous graph for events and users in logs using user-defined rules. LogKG\cite{b12} uses knowledge graph to establish association relationship between logs. Meanwhile, LogFlash\cite{b10} constructs a time series graph based on time parameters in logs, and detects log anomalies using time-based graph.

Current mainstream methods do not combine event templates and parameters for log anomaly detection. However, there may be inherent connections and rules between the event templates and parameters in logs. For instance, a log entry's template may indicate a normal operation type, but if its parameters are inconsistent with those of other logs or do not match the system's status, it may be considered as an abnormal log. On the other hand, a log event template may represent an unusual operation, but if its parameters are consistent with those of other logs or match the system's status at the time, it may be considered as a normal log. Therefore, detecting log anomalies can be improved by considering the event templates and parameters of logs as a whole.

\section{Design of TPLogAD}

\subsection{Overview}
\begin{figure}
    \centering
    \includegraphics[width=1\linewidth]{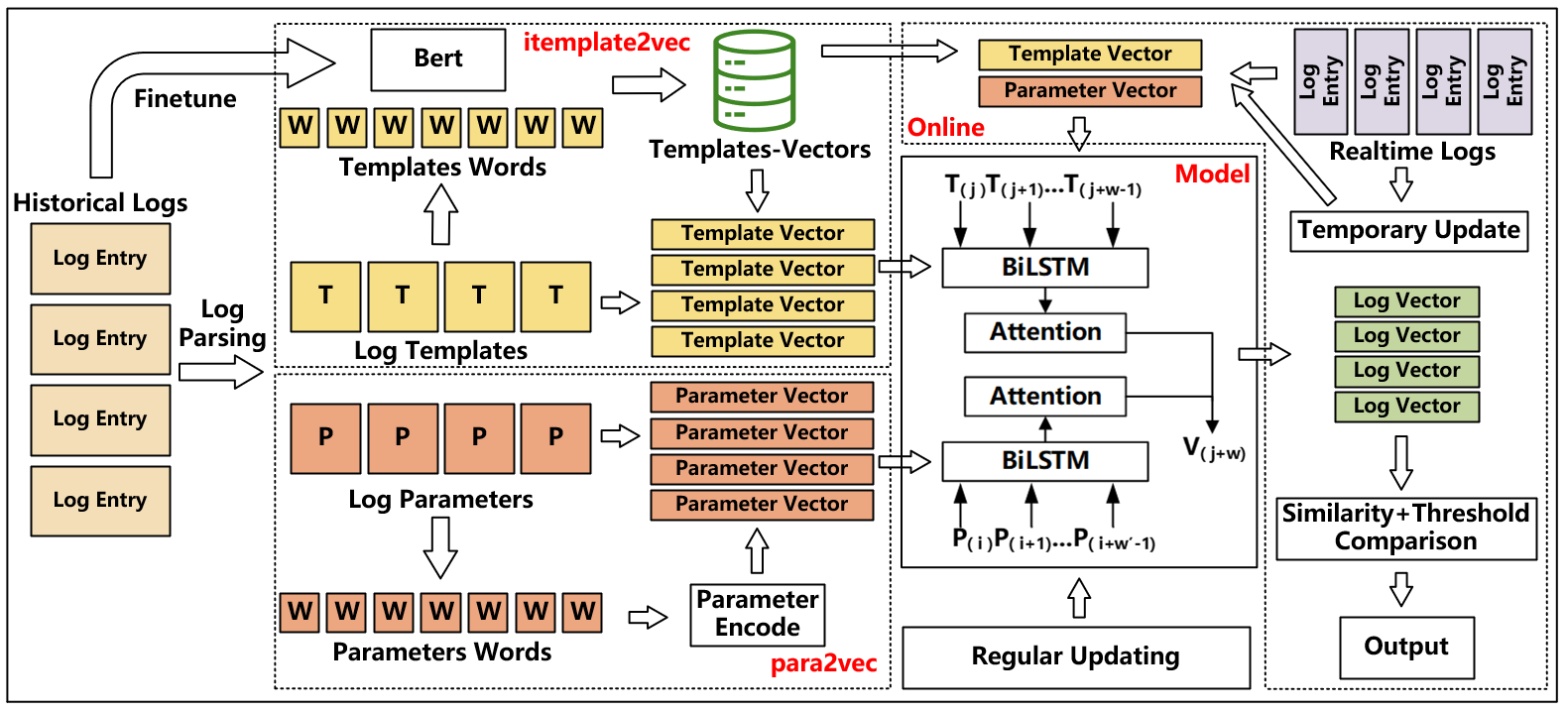}
    \caption{The framework of TPLogAD}
    \label{fig:TPLogAD}
\end{figure}
In the current reality, there are numerous log types and no unified standard for log generation has been established. Therefore, the industry prefer to use unsupervised log anomaly detection methods with high accuracy, which can automatically detect system failures from logs with minimal human intervention. The survey of real-world logs reveals that logs contain rich semantic and association information\cite{b4,b21}. Mainstream log parsing methods decompose log entries into templates and parameters. The templates contain the semantic information of the log entry, such as the action, result and cause of the event. These semantic details reflect the true meaning and purpose of the log entry recording a certain type of event. The parameters contain the event's association information, including the time, location, status, and attribute value of the event. This information may reflect the dynamic changes of the system status and imply rules and constrains of these changes under normal situation.

The main concept of our model is to utilize deep learning techniques to acquire normal semantic and association information from logs offline, and subsequently identify log anomalies online. Fig.~\ref{fig:TPLogAD} illustrates the structure of our proposed TPLogAD. During the left offline learning phase, Drain3 is employed to extract event templates and parameters from historical logs. For event templates, we propose itemplate2vec. The words of event templates are transformed into embedding vectors by semantic similarity, then template vectors are calculated by combining word vectors with semantic weights. For parameters, we propose para2vec. This method can extract the main parameters in the log and convert them into embedding vectors. Finally, the template vector and parameter vector are combined and input into the BiLSTM to learn the semantic and parameter representations of logs. To adapt to newly appeared representations, we have designed a method to characterize their similarity. The offline model is regularly updated to accommodate the diversity and dynamics of logs.

During the online detection process, TPLogAD receives real-time logs and converts them into template vectors and parameter vectors. If a real-time log entry does not have a matching template, TPLogAD selects the most similar template as a temporary template and extracts parameters. Hence, each real-time log entry can be transformed into a template vector and a parameter vector. TPLogAD subsequently compares these representations with the offline learned model. If the real-time logs' semantic representation and parameter representation deviate or mismatch from the offline learned model, it is considered abnormal.

\subsection{itemplate2vec}
\begin{figure}
    \centering
    \includegraphics[width=1\linewidth]{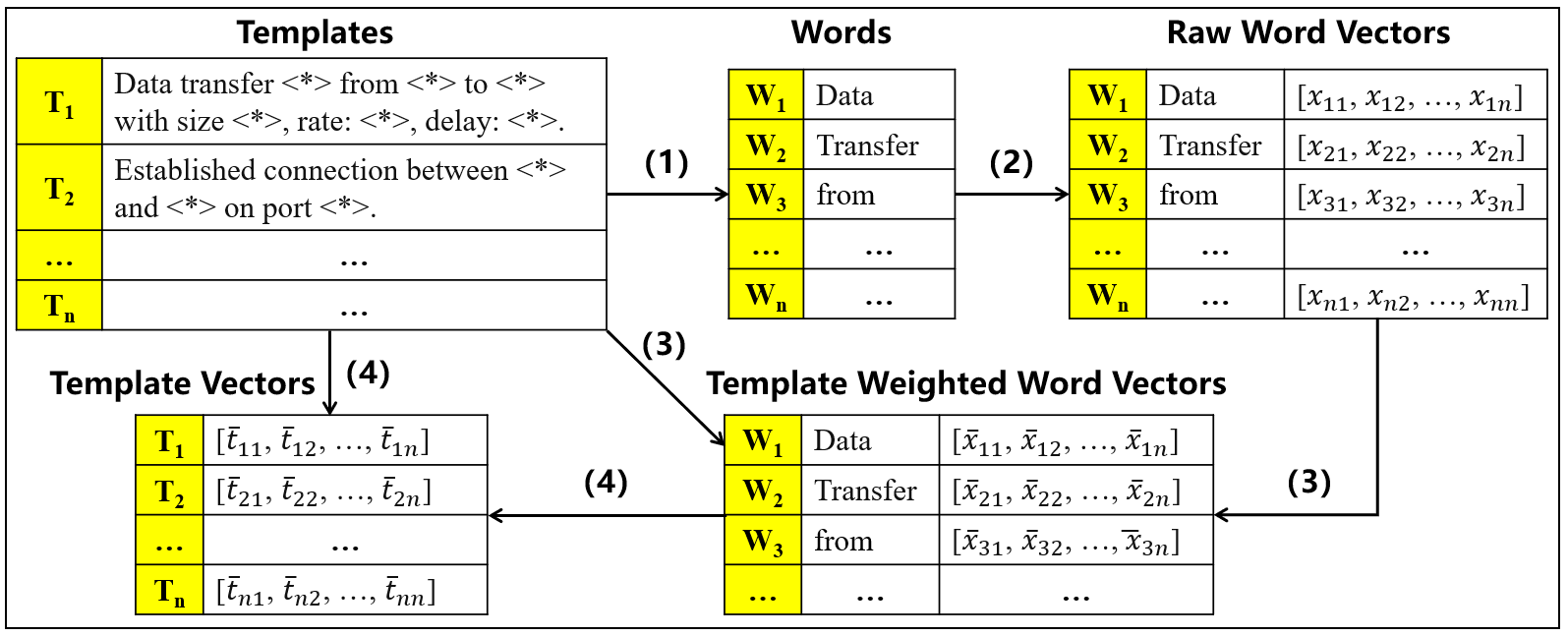}
    \caption{Examples of itemplate2vec}
    \label{fig:itemplate2vec}
\end{figure}
Log templates can reflect the semantic information of logs, including the action, object, result, and cause of the event. This information can help us understand and detect log anomalies. To represent templates as vectors, we require a distributed word representation method that reflects the contextual and semantic information of words in templates.

Word2vec\cite{b22} is a widely used word representation method. It can learn the distributed representation of words by using a large amount of text data, reflecting the contextual information of words. However, it cannot capture the semantic information of words. In contrast, template2vec\cite{b5} is a word representation method based on synonyms and antonyms, which can effectively distribute words in templates. It identifies synonyms and antonyms of words, then uses dLCE\cite{b23} to generate word vectors. However, it is unable to fully capture semantic information about words and can be poor at adapting to new words and patterns.
\begin{equation}
\lambda ({W_i}) = \frac{{\sum\limits_{j = 1,j \ne i}^n {\cos \_similarity({V_{old}}({W_i}),{V_{old}}({W_j}))} }}{{n - 1}}
\end{equation}
\begin{equation}
{V_{new}}({W_i}) = \lambda ({W_i}) \cdot {V_{old}}({W_i})
\end{equation}

To solve these issues, we propose a method for vectorizing templates based on semantic similarity, itemplate2vec, which can efficiently use the semantic information of the words in the template to construct template vectors. As shown in Fig.~\ref{fig:itemplate2vec}, itemplate2vec constructs template vectors in four steps during offline learning as follows: 

\begin{enumerate}[]
\item \textit{Create a log word table}. This involves traversing all log templates in a Web service system, treating each template word as an item, removing duplicates, and obtaining a log word table. 
\item \textit{Generate word vectors}. Bert\cite{b14} is a pre-trained language model that learns distributed word representation from large text data. The pre-trained Bert is fine-tuned using the historical logs. Then we use Bert's vector encoding results to generate a word vector for each word in the log word table. 
\item \textit{Calculate weighted word vectors}. Formula (1) is used to calculate the word's weight, where $n$ is the word's number of the log template and $V_{old}(W_i)$ is the original word vector from Bert. We use cosine similarity to measure the similarity between each word and other words in a log template. Then we use each word's average similarity to other words as the word's weight in the template. The weight reflects the proximity of a word to the semantic core in a template: the higher the weight, the closer it is to the semantic centre. Formula (2) is used to calculate the new word vector.
\item \textit{Calculate template vectors}. To represent the distribution of the template, the weighted average of the weighted word vectors in the template is calculated for each log template.
\end{enumerate}

During the online detection process, real-time logs are transformed into template vectors via itemplate2vec. If a matching template is found for the log, the existing template vector is used. Otherwise, the most similar template vector is selected as a temporary template vector.
\begin{figure}
    \centering
    \includegraphics[width=1\linewidth]{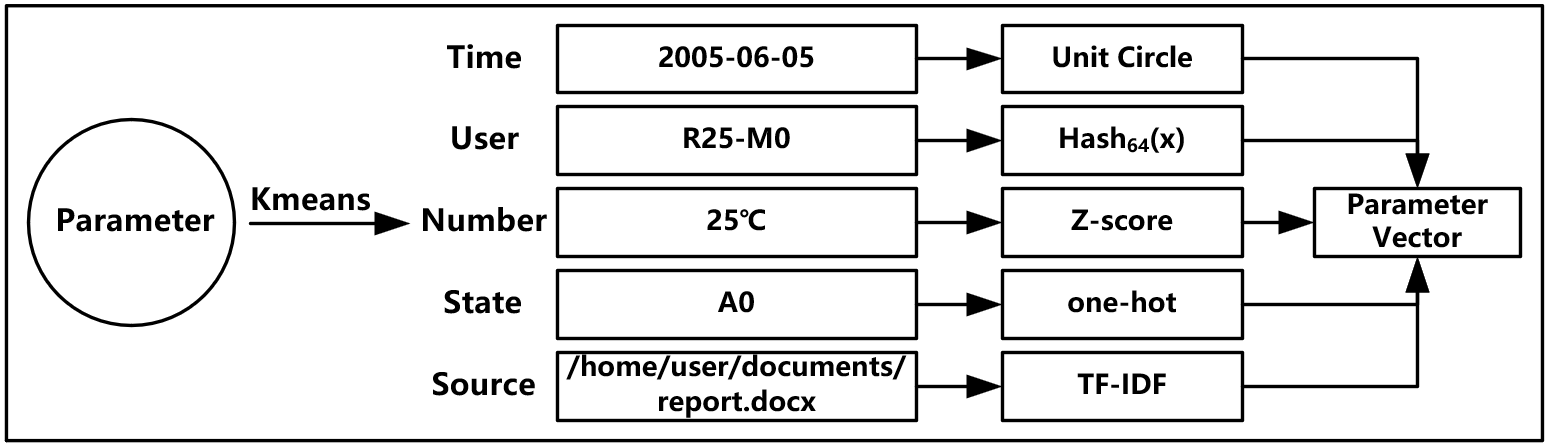}
    \caption{Examples of para2vec}
    \label{fig:para2vec}
\end{figure}

\subsection{para2vec}

Log parameters describe event properties and reflect the dynamic change rules of those properties. To create semantic and associative representations of logs that reflect the true meaning and purpose of logs, log parameters can be combined with the semantic information of event templates. For effective use of log parameter information, it is necessary to convert it into vector representations for anomaly detection, thus ensuring objectivity and precision in the analysis of the logs.

Time2vec\cite{b15} is a vector encoding method for time parameters. It uses a non-linear transformation based on sine and activation functions to map time parameters into vector space, ensuring that time parameters have similar vector representations while preserving periodicity of time parameters.

Inspired by time2vec, we propose para2vec, an encoding method for the five main parameter types commonly found in logs. As shown in Fig.~\ref{fig:para2vec}, the main parameters are first extracted from the parameter sequence via Kmeans clustering algorithm. Then the semantic and association information of different parameter types is converted into vector representations for log anomaly detection. The parameter types and their encoding methods are as follows:

\begin{enumerate}[]
    \item \textit{Time parameters}. These characterize the time of events and reflect the order, periodicity, and continuity of events. For example, they can indicate whether an event occurs at a fixed time point or not, or whether its number increases or decreases with time. Formula (3) is used to calculate the vector of time parameters, where $t$ is the time unit and $max_t$ is the maximum of that unit. For these parameters, cyclic distance encoding is applied to commonly occurring time units such as years, months, days, hours, minutes, seconds and milliseconds in logs, using their respective coordinates on the unit circle as vector representations. This approach ensures that closely spaced time parameters have similar vector representations, while preserving periodicity and continuity of time parameters, as well as relative distance and direction of them.
    \begin{equation}
    {V_{time}}(t) = (\sin (\frac{{2\pi t}}{{{{\max }_t}}}),\cos (\frac{{2\pi t}}{{{{\max }_t}}}))
    \end{equation}
    \item \textit{User identification parameters}. These parameters record the source of the event, which contains the user's personal information.  To identify the same subject that conducts a serial of malicious attacks and protect user's privacy, we use 64-bit hash encoding to encrypt the user identification parameters, ensuring that even if the encrypted parameters are leaked, adversaries cannot recover the original information. These parameters are finally normalized to a numerical vector within the same interval, eliminating scale differences between platforms or models. Formula (4) is used to calculate the user parameter vector, where $u$ is the user idnetification and $f_{hex\_to\_num}$ is the function that converts hexadecimal values to numbers.
    \begin{equation}
    {V_{user}} = {f_{hex\_to\_num}}(Has{h_{64}}(u))
    \end{equation}
    \item \textit{Numerical parameters}. These parameters refer to quantitative indicators of an event that reflect its performance and impact. Z-score normalisation is used to encode these parameters, which can be used to determine whether they meet expectations or have a mutation or abnormal value.  This method eliminates the influence of data dimensions and ranges, allowing comparison and analysis on the same scale. It also preserves the scale characteristics of numerical parameters and improves the stability and reliability of the data, while ensuring that abnormal values is not affected by extreme values.
    \item \textit{State parameters}. These parameters refer to discrete values that record the state of a certain feature in an event. They reflect the discrete state of a particular aspect of the event, such as whether the event was successfully completed, or whether a particular port was disconnected or connected. To normalize the state parameters into a numerical value, we use a transformation function based on one-hot encoding. Then we treat the encoded result as binary and convert it to a numeric value. Formula (5) is used to calculate the state parameter vector, where $s$ is the state, $f_{vec\_to\_bin}$ is the function that splice one-hot encoding by bits and $f_{bin\_to\_num}$ is the function that converts binary values to numbers.This enables state parameters to have distinct vector representations while preserving their categorical features.
    \begin{equation}
    {V_{state}} = {f_{bin\_to\_num}}({f_{vec\_to\_bin}}({f_{one\_hot}}(s))
    \end{equation}
    \item \textit{Resource identification parameters}. These parameters record the object of the event, such as the destination address, url or file path. We use a transformation based on word embedding to vectorize these parameters via TF-IDF\cite{b24}. Compared to the one-hot coding used in the past, this method reduces vector length and redundancy, improves vector discriminability, and simplifies similarity calculations. By measuring vector similarity in distance or angle, it can quickly identify similar or different resources in logs, as well as the hierarchical relationship between different resource identifications. Meanwhile, the semantic features of the resource identification parameters are preserved.
\end{enumerate}

Compared to traditional log anomaly detection methods based on clustering, para2vec identifies abnormal logs more accurately by capturing inherent log parameter characteristics and trends. Traditional methods typically overlook the type and information of parameters and only process logs as text, resulting in information loss and error accumulation. Para2vec can convert parameters into vector representations and preserve the original information while increasing their computability. This transforms log analysis and anomaly detection into classic vector operations and machine learning problems, improving effectiveness and efficiency of log anomaly detection.

\subsection{Log Anomaly Detection Based on Template And Parameter}

\begin{figure}
    \centering
    \includegraphics[width=0.8\linewidth]{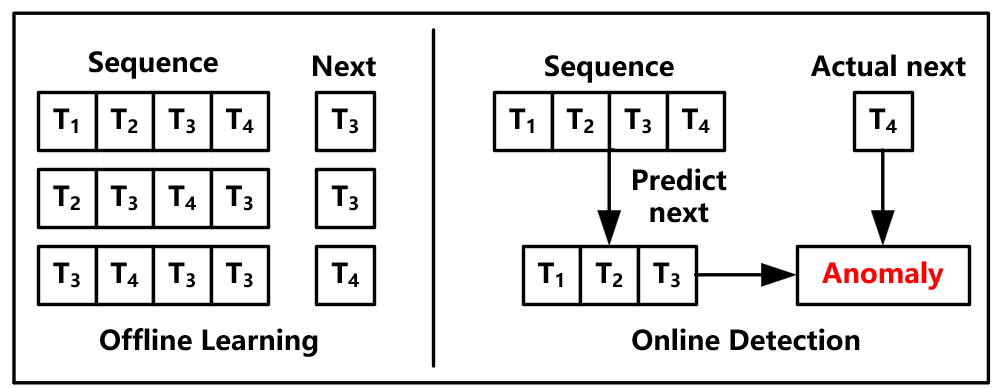}
    \caption{Examples of Template Anomaly}
    \label{fig:template_anomaly}
\end{figure}
\begin{figure}
    \centering
    \includegraphics[width=0.8\linewidth]{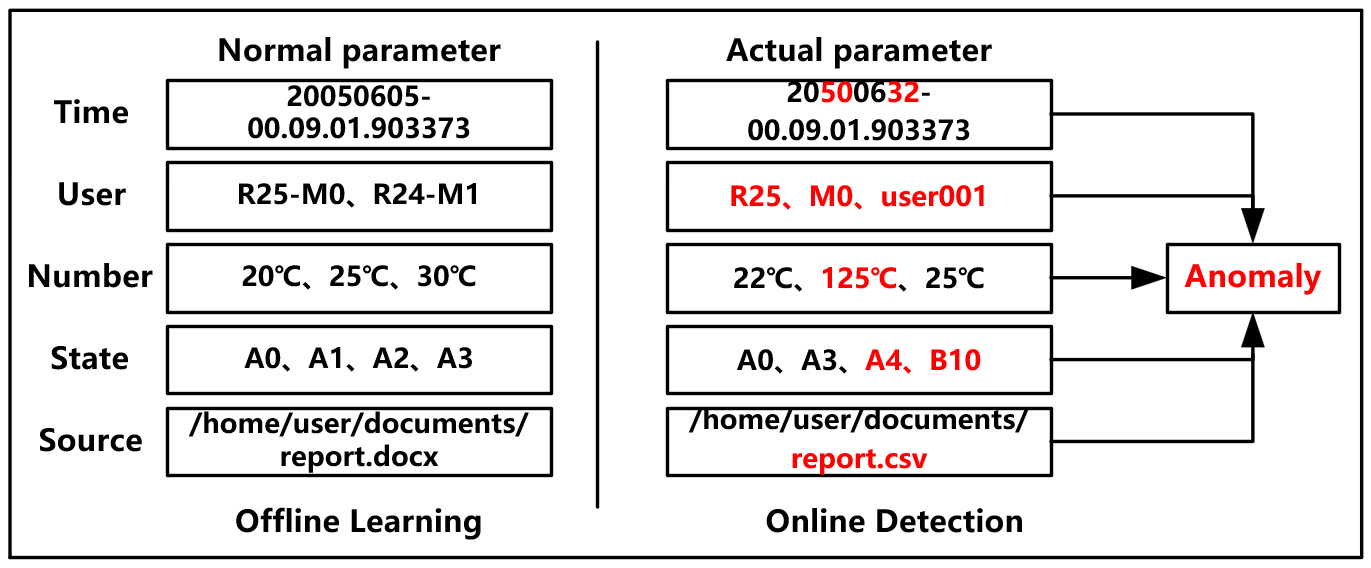}
    \caption{Examples of Parameter Anomaly}
    \label{fig:parameter_anomaly}
\end{figure}
Developers' log statements are executed in a specific order within the program, resulting in identifiable patterns and associations between them. To capture this, we convert the log sequence into template and parameter vectors using itemplate2vec and para2vec. These vectors are then inputted into a BiLSTM with an attention mechanism to learn the semantic and associative information of the logs.

Assuming the current complete log stream is $S=(L_1,L_2,...,L_n)$, we obtain the template sequence $TS=(T_1,T_2,...,T_n)$ and the parameter sequence $PS = (P_1,P_2,...,P_n)$ after parsing it. For the template sequence of the log, the detection window is $w$ log entries. Each time, we take out $w$ log entries from the log stream as a detection unit and convert each log's template into a vector representation. Suppose $TS_j=(T_j,T_{j + 1},...,T_{j + w - 1})$ is one of the detection windows. Fig.~\ref{fig:template_anomaly} shows that the template sequence in the normal log stream follows a specific order and has a certain association. The objective is to predict the next log template in the detection window. If the actual template is not in the predicted candidate sequence, it is considered as an abnormal template sequence.

For the log parameter sequence, the detection window is $w'$ log entries. Each detection unit consists of $w'$ log entries with the same event template from the log stream, and each entry's parameters are converted to vector representations. If a log entry contains multiple parameters, the vector representations of each parameter are merged according to their types to form the log entry's parameter vector. Assume that $PS_i=(P_i,P_{i + 1},...,P_{i + w' - 1})$ is one of the detected windows. Based on extensive research, the five major parameter types have different anomaly judging criteria, as shown in Fig.~\ref{fig:parameter_anomaly}. For time parameters, if their format does not conform to the standard, or exceed the reasonable range, the time order should be considered as incorrect. For user identification parameters, if their value is empty, or if the encoding result is not consistent with the majority of user identifications, then it is considered to be an abnormal parameter. For numerical parameters, an abnormal parameter is one whose value is not a valid number or is outside the reasonable range. For state parameters, an abnormal parameter is one that does not belong to the predefined set of enumerations or that changes too frequently. For resource identification parameters, an abnormal parameter is one whose value is an incorrect identification path or whose association with other parameters of the same type is low.

To summarise, TPLogAD converts the template and parameter sequence into vectors to detect anomalies, using various window sizes. The order of logs in the window and the association between vectors are the criteria for judging anomalies, while different types of parameters in the window are analyzed to identify anomalies in the parameter sequence. The experimental results in Section IV demonstrate that our method can effectively utilize semantic and association information in logs. By comprehensively considering anomalies in the template and parameter sequence, it can more accurately identify abnormal logs, thereby improving the effectiveness and efficiency of log anomaly detection.

\subsection{Update of Template and Parameter}

Web service systems require an anomaly detection method that can adapt to dynamic changes in logs. Existing methods cannot handle newly generated templates and parameters online. Our approach is to reduce feedbacks from administrators on new templates and parameters, minimising labour costs and avoiding delays.

During the online detection phase, we extract event templates and parameters from each log entry using Drain3. If no matching event template is found, we calculate a new template vector using itemplate2vec. Finally, the temporary template vector is found by identifying the existing template vector with the highest similarity to this new template vector. Our method has the advantage that it does not require any manual feedback from the operators or frequent offline retraining, and it can also deal effectively with newly appearing event templates and parameters. In this way, our method is able to adapt to the dynamic changes of the logs in real-time systems, thereby improving the accuracy and efficiency of the anomaly detection.

\begin{table}[ht!]
    \centering
    \caption{Detail of the log datasets}
    \begin{tabular}{ccc}
        \toprule
        Log datasets & Number of log & Number of anomaly \\
        \midrule
        BGL         & 4,747,963   & 348,460 \\
        HDFS        & 11,175,629  & 16,838 \\
        ThunderBird & 211,212,192 & 3,248,239 \\
        Spirit      & 272,298,969 & 172,816,564 \\
        \bottomrule
    \end{tabular}
    \label{tab:log_datasets}
\end{table}

\section{Evaluation}

\subsection{Experiment Setting}

As shown in Table \ref{tab:log_datasets}, the experiments are performed on four public log datasets: BGL, HDFS, ThunderBird and Spirit\cite{b21,b25}. BGL refers to BlueGene/L supercomputer log, HDFS refers to Hadoop distributed file system log, ThunderBird refers to ThunderBird supercomputer log and Spirit refers to Spirit supercomputer log. The BGL and ThunderBird datasets have fewer log events, but the logs are highly correlated. The remaining two datasets have more events logged but are weakly correlated. For each log dataset in the following experiments, we use 80\% as the training data, and the rest 20\% as the testing data.

The TPLogAD method proposed in this paper is compared with DeepLog\cite{b8}, LogAnomaly\cite{b5}, Logsy\cite{b7}, and LogBERT\cite{b20}, which have been shown in previous research to achieve the best results in log anomaly detection. We use Precision, Recall, and F1-score as performance measurement indicators, and mark the predicted results as TP, TN, FP, and FN. TP is the abnormal log correctly identified as abnormal, TN is the normal log correctly identified as normal, FP is the normal log incorrectly identified as abnormal and FN is the abnormal log incorrectly identified as normal. Precision = $\frac{TP}{(TP+FP)}$, Recall = $\frac{TP}{(TP+FN)}$, F1-score = $\frac{2*Percision*Recall}{Percision+Recall}$. Precision measures the proportion of true anomalies among the detected anomalies, while recall measures the proportion of true anomalies that are detected. The F1-score is the harmonic mean of accuracy and recall, providing an overall measure of the model's performance.

\subsection{Evaluation of The Overall Performance}

In this section, we compare the performance of TPLogAD and the four baseline methods on four publicly available log datasets. Our experiments use a BiLSTM neural network with 256 neurons, with a window size of 20 for template sequence and 100 for parameter sequence. And we use \textit{bert-base-uncased} as the pre-trained model of Bert.

Table \ref{tab:acc-BTHS} shows that TPLogAD performs best on all datasets, with average F1-scores of 0.95 and 0.96 for the BGL and ThunderBird datasets, and 0.95 and 0.97 for the HDFS and Spirit datasets. In contrast, the F1-scores of DeepLog and LogBERT vary widely across the different datasets, ranging from 0.60 to 0.91. LogAnomaly has an F1-score of 0.92 on both the BGL and HDFS datasets, but this drops to 0.83 on the Spirit dataset. Logsy's performance is relatively stable, but not as high as TPLogAD.

\begin{table*}[ht!]
    \centering
    \caption{Accuracy on four log datasets(BGL, ThunderBird, HDFS and Spirit)}
    \begin{tabular}{ccccccccccccc}
        \toprule
        \multirow{2}{*}{Method} & \multicolumn{3}{c}{BGL} & \multicolumn{3}{c}{ThunderBird} & \multicolumn{3}{c}{HDFS} & \multicolumn{3}{c}{Spirit}\\
        \cmidrule(lr){2-4} \cmidrule(lr){5-7} \cmidrule(lr){8-10} \cmidrule(lr){11-13}
        & Pre & Rec & F1 & Pre & Rec & F1 & Pre & Rec & F1 & Pre & Rec & F1\\
        \midrule
        DeepLog & 0.88 & 0.93 & 0.90 & 0.83 & 0.87 & 0.85 & 0.92 & 0.90 & 0.91 & 0.57 & 0.63 & 0.60\\
        LogAnomaly & 0.94 & 0.91 & 0.92 & 0.88 & 0.91 & 0.89 & 0.93 & 0.92 & 0.92 & 0.81 & 0.85 & 0.83\\
        Logsy & 0.86 & 0.93 & 0.89 & 0.93 & 0.92 & 0.92 & 0.87 & 0.89 & 0.88 & 0.92 & 0.89 & 0.90\\
        LogBERT & 0.89 & 0.92 & 0.90 & 0.92 & 0.91 & 0.91 & 0.88 & 0.83 & 0.85 & 0.78 & 0.75 & 0.76\\
        TPLogAD w/o itemplate2vec & 0.92 & 0.93 & 0.92 & 0.91 & 0.92 & 0.91 & 0.93 & 0.92 & 0.92 & 0.94 & 0.95 & 0.94\\
        TPLogAD w/o para2vec & 0.95 & 0.92 & 0.93 & 0.96 & 0.95 & 0.95 & 0.92 & 0.96 & 0.94 & 0.95 & 0.96 & 0.95\\
        TPLogAD & \textbf{0.96} & \textbf{0.95} & \textbf{0.95} & \textbf{0.97} & \textbf{0.96} & \textbf{0.96} & \textbf{0.94} & \textbf{0.97} & \textbf{0.95} & \textbf{0.98} & \textbf{0.97} & \textbf{0.97}\\
        \bottomrule
    \end{tabular}
    \label{tab:acc-BTHS}
\end{table*}

\begin{table*}[ht!]
    \centering
    \caption{Accuracy on different divisions of training dataset in online anomaly detection}
    \begin{tabular}{cccccccccccccc}
        \toprule
        \multirow{2}{*}{Dataset Division} & \multirow{2}{*}{Method} & \multicolumn{3}{c}{BGL} & \multicolumn{3}{c}{ThunderBird} & \multicolumn{3}{c}{HDFS} & \multicolumn{3}{c}{Spirit}\\
        \cmidrule(lr){3-5} \cmidrule(lr){6-8} \cmidrule(lr){9-11} \cmidrule(lr){12-14}
        & & Pre & Rec & F1 & Pre & Rec & F1  & Pre & Rec & F1 & Pre & Rec & F1\\
        \midrule
        & DeepLog & 0.81 & 0.82 & 0.81 & 0.80 & 0.81 & 0.80 & 0.85 & 0.83 & 0.84 & 0.46 & 0.53 & 0.49\\
        & LogAnomaly & 0.89 & 0.87 & 0.88 & 0.86 & 0.88 & 0.87 & 0.87 & 0.84 & 0.85 & 0.71 & 0.75 & 0.72\\
        60\% Dataset & Logsy & 0.83 & 0.88 & 0.85 & 0.83 & 0.84 & 0.83 & 0.82 & 0.83 & 0.82 & 0.87 & 0.85 & 0.86\\
        & LogBERT & 0.85 & 0.88 & 0.86 & 0.81 & 0.77 & 0.79 & 0.80 & 0.77 & 0.78 & 0.72 & 0.71 & 0.71\\
        & TPLogAD & \textbf{0.93} & \textbf{0.91} & \textbf{0.92} & \textbf{0.92} & \textbf{0.95} & \textbf{0.93} & \textbf{0.92} & \textbf{0.89} & \textbf{0.90} & \textbf{0.95} & \textbf{0.93} & \textbf{0.94}\\
        \midrule
        & DeepLog & 0.74 & 0.69 & 0.71 & 0.67 & 0.72 & 0.69 & 0.72 & 0.65 & 0.68 & 0.33 & 0.36 &0.34\\
        & LogAnomaly & 0.80 & 0.81 & 0.80 & 0.77 & 0.76 & 0.76 & 0.79 & 0.78 & 0.78 & 0.63 & 0.66 &0.64\\
        50\% Dataset & Logsy & 0.72 & 0.63 & 0.67 & 0.74 & 0.79 & 0.76 & 0.72 & 0.75 & 0.73 & 0.70 & 0.71 &0.70\\
        & LogBERT & 0.65 & 0.59 & 0.62 & 0.62 & 0.67 & 0.64 & 0.63 & 0.62 & 0.62 & 0.59 & 0.60 & 0.59\\
        & TPLogAD & \textbf{0.85} & \textbf{0.89} & \textbf{0.87} & \textbf{0.88} & \textbf{0.92} & \textbf{0.89} & \textbf{0.87} & \textbf{0.84} & \textbf{0.85} & \textbf{0.88} & \textbf{0.87} & \textbf{0.87}\\
        \midrule
        & DeepLog & 0.39 & 0.36 & 0.37 & 0.41 & 0.37 & 0.39 & 0.42 & 0.35 & 0.38 & 0.29 & 0.32 &0.30\\
        & LogAnomaly & 0.43 & 0.45 & 0.44 & 0.51 & 0.49 & 0.50 & 0.46 & 0.43 & 0.44 & 0.41 & 0.38 &0.39\\
        40\% Dataset & Logsy & 0.43 & 0.49 & 0.46 & 0.62 & 0.53 & 0.57 & 0.48 & 0.45 & 0.46 & 0.43 & 0.44 & 0.43\\
        & LogBERT & 0.45 & 0.41 & 0.43 & 0.54 & 0.49 & 0.51 & 0.39 & 0.36 & 0.37 & 0.35 & 0.32 & 0.33\\
        & TPLogAD & \textbf{0.82} & \textbf{0.85} & \textbf{0.83} & \textbf{0.84} & \textbf{0.89} & \textbf{0.86} & \textbf{0.79} & \textbf{0.83} & \textbf{0.81} & \textbf{0.81} & \textbf{0.80} & \textbf{0.80}\\
        \bottomrule
    \end{tabular}
    \label{tab:acc-654}
\end{table*}

To illustrate the advantages of merging template vectors and parameter vectors, we calculate the accuracy of TPLoqAD without(w/o) itemplate2vec and without(w/o) para2vec. Table \ref{tab:acc-BTHS} demonstrates that TPLoqAD's precision is significantly lower on the BGL dataset (0.92) without itemplate2vec, compared to the full TPLoqAD. And the precision also decreases on the ThunderBird dataset (0.91). Thus, TPLoqAD's itemplate2vec prevents false positives from interfering with administrators. Similarly, para2vec also improves both precision and recall for TPLogAD on the four datasets.

As a log anomaly detection method based on semantic extraction of event templates and parameter correlation analysis, TPLogAD achieves excellent results on various datasets. It can effectively capture semantic information and parameter information in logs, and has high robustness and generalization ability. TPLogAD achieves high precision and recall on all four log datasets. For instance, on the BGL dataset, TPLogAD achieves a precision of 0.96 and a recall of 0.95. This indicates that TPLogAD can accurately detect abnormal logs across various scenarios without producing excessive false alarms. Given that Web service systems generate millions of logs daily, a log anomaly detection method that generates too many false alarms can significantly burden administrators. TPLogAD utilizes itemplate2vec to learn the semantic information of log templates, rather than relying solely on template indexes or the semantic information of synonyms and antonyms to distinguish between abnormal and normal patterns. This approach enables TPLogAD to handle different template indexes, while also enhancing the semantic association between different logs. Furthermore, TPLogAD utilizes para2vec to comprehend the semantic information and association of parameters in logs, instead of disregarding the parameter section of logs like other methods. This enables TPLogAD to handle parameter changes in logs, such as the parameters highlighted in red in Fig.~\ref{fig:log-example}, which can provide significant information about log events. Previous methods only considere event types indicated by templates in logs and thus discard this information. By utilizing this parameter information, TPLogAD can detect abnormal logs more accurately than other methods.

\subsection{Evaluation of Online Anomaly Detection}

Web service systems continuously generate new types of logs, including new log templates and parameters due to frequent software or system updates that bring new features or fix old errors. An anomaly detection method that cannot adapt to new changes will generate false alarms. This is because the new changes cause log templates and parameters to mismatch with previous normal patterns.

For the BGL dataset, assume that the training set is 60\% of the dataset. 60\% of the data contain 602 log templates, and the rest 40\% contain 206 log templates. The training sets of the BGL dataset will consist of 602 log templates. However, it is important to note that approximately 12.6\% of the test data, which equates to 239,338 log entries, will cannot be matched to any log template. Similarly, the situation for the other log datasets is consistent with the BGL dataset, reflecting the real-world problem of logs not matching due to updates in Web service systems.

To evaluate the effectiveness of TPLogAD in handling the challenges posed by the dynamic and diverse nature of logs, we train the model on 60\%, 50\%, and 40\% of the four log datasets respectively, and use the rest 40\%, 50\%, and 60\% of the data for testing. The training datasets are divided into three cases: 60\% represents the scenario where the system is less updated and fewer logs cannot be matched, 50\% represents the moderately updated scenario, and 40\% represents the frequently updated scenario where a large number of logs cannot be matched. We conduct comparative experiments to confirm that we can still achieve the best results in complex and variable environments.

The experimental results presented in Table \ref{tab:acc-654} show that TPLogAD performs exceptionally well across all training set divisions of the two datasets. Specifically, when using 60\%, 50\%, and 40\% of the data as training sets on the BGL dataset, TPLogAD achieves F1-scores of 0.92, 0.87, and 0.83, which are significantly higher than those achieved by other methods. Similarly, TPLogAD also demonstrates outstanding performance on other datasets.

In contrast, Logsy and LogBERT performed well but cannot outperform TPLogAD in precision, recall and F1-score. Additionally, their performance sharply decline with the reduction of the training set. For example, using only 40\% of the training data on the Spirit dataset, the F1-scores of DeepLog, LogAnomaly, Logsy and LogBERT are only 0.30, 0.39, 0.43 and 0.33 respectively, while TPLogAD achieves 0.80. These results show that TPLogAD has a significant advantage in online detection, and its stability and accuracy are superior to current state-of-the-art methods.

\section{Conclusion}

Logs are a kind of valuable data sources for recording the operational status and behaviour of Web service systems. Analyzing logs can help relevant personnels to identify faults and security issues. In this paper, we propose TPLogAD, an unsupervised log anomaly detection based on event templates and key parameters. This method utilizes itemplate2vec to extract the semantic information of event templates and para2vec to extract the associative information of key parameters. Numerous experiments have been shown that TPLogAD able to effectively extract semantic and associative information from logs in various environments, and to identify abnormal logs.

\vspace{12pt}

\end{document}